\documentclass[runningheads]{llncs}

\usepackage{epsfig}
\usepackage{graphicx}
\usepackage{amsmath,amssymb}
\usepackage[nowatermark]{fixmetodonotes}
\usepackage[utf8]{inputenc} 
\usepackage[T1]{fontenc}    
\usepackage{url}            
\usepackage{booktabs}       
\usepackage{amsfonts}       
\usepackage{nicefrac}       
\usepackage{enumitem}	
\graphicspath{
{../Figures/}
}

\def \FPT{{\hyperref[FPTDef]{FPT}}}
\def \UFPT{{\hyperref[UFPTDef]{{\sc UFPT}}}}
\def \NPT{{\hyperref[NPTDef]{{\sc NPT}}}}

\usepackage[pagebackref=true,breaklinks=true,bookmarks=false]{hyperref}

\usepackage{microtype}

\begin{document}
\title{Whole MILC: generalizing learned dynamics across tasks, datasets, and populations}
\titlerunning{Whole MILC}
%
\author{Usman Mahmood\inst{1} \and
Md Mahfuzur Rahman\inst{1} \and
Alex Fedorov \inst{2} \and
Noah Lewis \inst{2} \and
Zening Fu \inst{1} \and
Vince D. Calhoun \inst{1,2,3} \and
Sergey M. Plis\inst{1}}

%
\authorrunning{U. Mahmood et al.}
%
\institute{Georgia State University\\ \email{\{umahmood1,mrahman21\}@student.gsu.edu}, \email{\{zfu,vcalhoun,splis\}@gsu.edu}\and Georgia Institute of
  Technology\\\email{{afedorov}@gatech.edu}, \email{{lhd231}@gmail.com} \and
  Emory University\\
  Atlanta, GA, USA \\
}
\maketitle              
%



\begin{abstract}
Behavioral changes are the earliest signs of a mental disorder, but arguably, the dynamics of brain function gets affected even earlier.  Subsequently, spatio-temporal structure of disorder-specific dynamics is crucial for early diagnosis and understanding the disorder mechanism.  A common way of learning discriminatory features relies on training a classifier and evaluating feature importance.  Classical classifiers, based on handcrafted features are quite powerful, but suffer the curse of dimensionality when applied to large input dimensions of spatio-temporal data.  Deep learning algorithms could handle the problem and a model introspection could highlight discriminatory spatio-temporal regions but need way more samples to train.  In this paper we present a novel self supervised training schema which reinforces whole sequence mutual information local to context (whole MILC).  We pre-train the whole MILC model on unlabeled and unrelated healthy control data.  We test our model on three different disorders (i) Schizophrenia (ii) Autism and (iii) Alzheimers and four different studies.  Our algorithm outperforms existing self-supervised pre-training methods and provides competitive classification results to classical machine learning algorithms.  Importantly, whole MILC enables attribution of subject diagnosis to specific spatio-temporal regions in the fMRI signal.


 \end{abstract}

\section{Introduction}
Mental disorders manifest in behavior that is driven by disruptions in brain dynamics~\cite{goldberg1992common,calhoun2014chronnectome}.
Functional MRI captures the nuances of spatio-temporal dynamics that could potentially provide clues to the causes of mental disorders and enable early diagnosis.
However, the obtained data for a single subject is of high dimensionality $m$ and to be useful for learning, and statistical analysis, one needs to collect datasets with a large number of subjects $n$.
Yet, for any kind of a disorder, demographics or other types of conditions, a single study is rarely able to amass datasets large enough to go out of the $m\gg n$ mode.
Traditionally small data problem is approached by handcrafting features~\cite{Khazaee2016} of much smaller dimension, effectively reducing $m$ via dimensionality reduction.
Often, the dynamics of brain function in these representations vanishes into proxy features such as correlation matrices of functional network connectivity (FNC)~\cite{yan2017discriminating}.

Our goal is to enable the direct study of brain dynamics in the $m\gg n$ situation. In the case of brain data it, in turn, can enable an analysis of brain function via model introspection. In this paper, we show how one can achieve significant improvement in classification directly from dynamical data on small datasets by taking advantage of publicly available large but unrelated datasets.
We demonstrate that it is possible to train a model in a self-supervised manner on dynamics of healthy control subjects from the Human Connectome Project (HCP)~\cite{van2013wu} and apply the pre-trained model to a completely different data collected across multiple sites from healthy controls and patients. We show that pre-training on dynamics allows the encoder to generalize across a number of datasets and a wide range of disorders: schizophrenia, autism, and Alzheimer's disease. Importantly, we show that learnt dynamics generalizes across different data distributions, as our model pre-trained on healthy adults shows improvements in children and elderly.

\section{Related Work}
Unsupervised pre-training is a well-known technique to get a head start for the deep neural network~\cite{erhan2010does}. It finds wide use across a number of fields such as computer vision~\cite{henaff2019data}, natural language processing (NLP)~\cite{devlin2018bert} and automatic speech recognition (ASR)~\cite{lugosch2019speech}. However, outside NLP unsupervised pre-training is not as popular as supervised.

Recent advances in self-supervised methods with mutual information objectives are approaching performance of supervised training~\cite{infonce,hjelm2018learning,bachman2019learning} and can scale pre-training to very deep convolutional networks (e.g., 50-layer ResNet). They were shown to benefit structural MRI analysis~\cite{fedorov2019prediction}, learn useful representations from the frames in Atari games~\cite{anand2019unsupervised} and for speaker identification~\cite{ravanelli2018learning}. Pre-trained models can outperform supervised methods by a large margin in case of small data~\cite{henaff2019data}.

Earlier work in brain imaging~\cite{khosla2019machine,frontiers2014} have been based on unsupervised methods to learn the dynamics and structure of the brain using approaches such as ICA~\cite{calhoun2001method} and HMM~\cite{eavani2013unsupervised}.
Deep learning for capturing the brain dynamics has also been previously proposed~\cite{hjelm2014restricted,hjelm2018spatio,khosla2019detecting}.
In some very small datasets, transfer learning was proposed for use in neuroimaging applications~\cite{mensch2017learning,10.3389/fnins.2018.00491,thomas2019deep}.
Yet another idea is the data generating approach~\cite{ulloa2018improving}. ST-DIM~\cite{anand2019unsupervised} has been used for pre-training on unrelated data with subsequent use for classification~\cite{mahmood2019transfer}.

\section{MILC}
\label{milc}
We present MILC as an unsupervised pre-training method. We use MILC to pre-train on large unrelated and unlabelled data to better learn data representation. The learnt representations are then used for classification on downstream tasks adding a simple linear network on top of the pre-training architecture.
The fundamental idea of MILC is to establish relationship between windows (a time slice from the entire sequence) and their respective sequences through learning useful signal dynamics. In all of our experiments we use encoded rsfMRI ICA time courses as our sequences and a consecutive chunk of time points as windows. The model uses the idea to distinguish among sequences (subjects) which proves to be extremely useful in downstream tasks e.g classification of HC or SZ subjects.
To realize the concept, we maximize the mutual information of the latent space of a window and the corresponding sequence as a whole.

Let $D = \{(u_t^i, {v}^j): 1 \le t \le T, 1 \le i,j \le N \}$ be a dataset of pairs computed from ICA time courses. $u_t^i$ is the local embedding of $t$-th window taken from sequence $i$, ${v}^j$ is the global embedding for the entire sequence $j$. $T$ is the number of windows in a sequence, and $N$ is the total number of sequences. Then $D^+ = \{(u_t^i, {v}^j ):  1 \le t \le T, i = j \}$ is called a dataset of positive pairs and $D^- = \{(u_t^i, {v}^j ):  1 \le t \le T, i \ne j \}$ --- of negative pairs. The dataset $D^+$ refers to a joint distribution and $D^-$ --- a marginal distribution of the whole sequence and the window in the latent space. Eventually, the lower bound with InfoNCE estimator~\cite{infonce} $\mathcal{I}_f(D^+)$ is defined as:

\begin{equation}
 \mathcal{I}(D^+) \ge \mathcal{I}_f(D^+) \triangleq \sum^N_{i=1} \sum^T_{t=1} \log \frac{\exp f((u_t^i, {v}^i))}{\sum^N_{k=1} \exp f((u_t^i, {v}^k))},
\end{equation}

where $f$ is a critic function. Specifically, we are using separable critic $f(u_t,v_s) = \phi(u_t^i)^\intercal (v^j)$, where $\phi$ is some embedding function parameterized by neural networks. Such embedding function is used to calculate value of a critic function in same dimensional space from two dimensional inputs. Critic learns an embedding function such that critic assigns higher values for positive pairs compared to negative pairs: $f(D^+) \gg f(D^-)$.

\begin{figure}
  \includegraphics[height=7cm ]{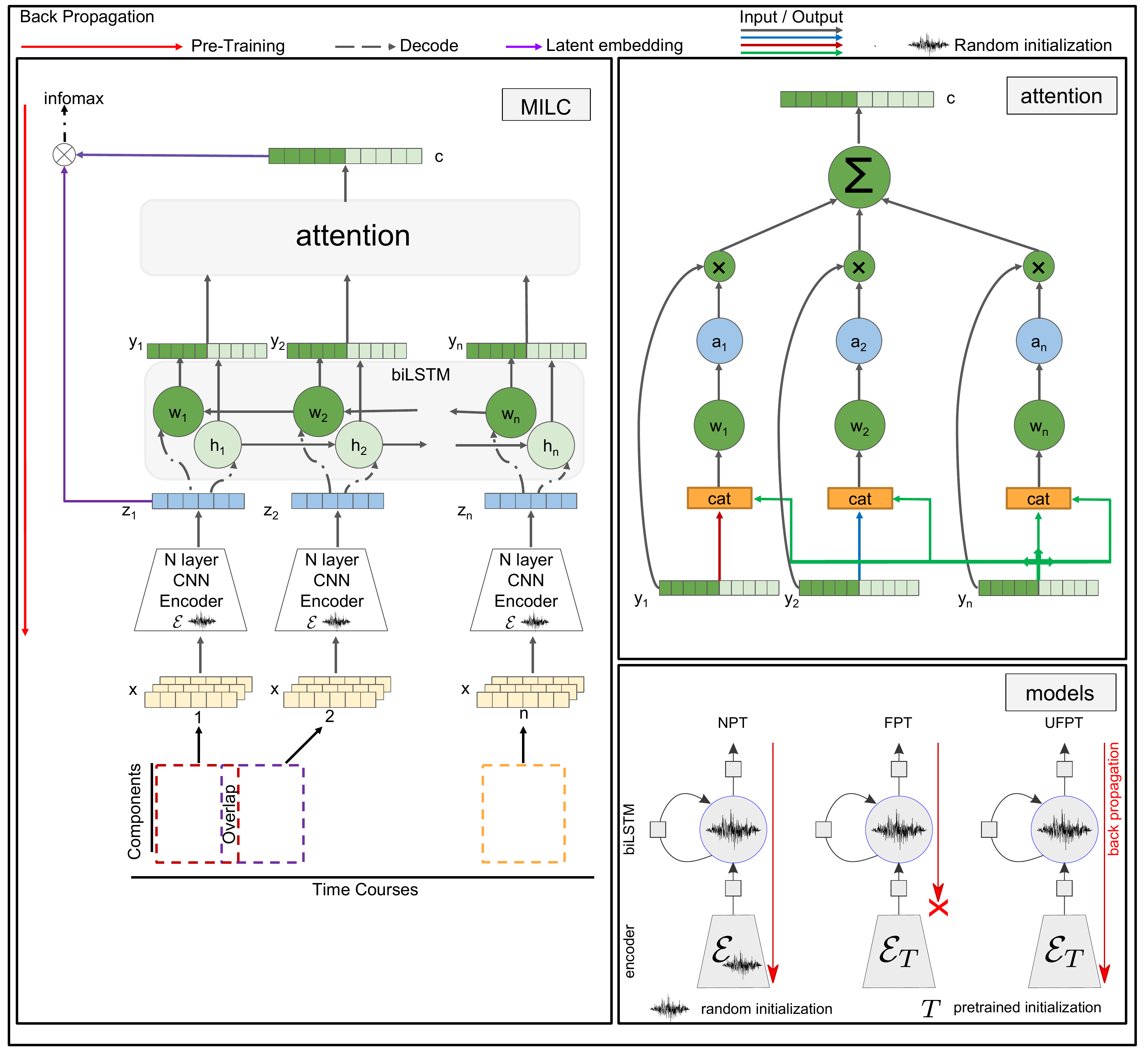}
  \centering
  \caption{\textbf{Left:} MILC architecture used in pre-training. ICA time courses are computed from the rsfMRI data. Results contain statistically independent spatial maps (top) and their corresponding time courses. \textbf{Right Up:} Detail of attention model used in MILC. \textbf{Right Down:} Three different models are used for downstream tasks.}
  \label{fig:milc}
\end{figure}

Our critic function takes the latent representation of a window and sequence as input. We define latent state of window as an output $z_t^i$ produced by the CNN part of MILC, given input from $t$-th window $x_t^i$ of sequence $i$. The latent state of sequence as $c^j$ is the global embedding obtained from MILC architecture. Thus the critic function for input pair $(x_t^i, x^j)$---a window and a sequence---is $f = \phi(z_t^i)^\intercal (c^j)$. The loss is InfoNCE with $f$ as $L = - \mathcal{I}_{f}$. The scheme of the MILC is shown in Figure~\ref{fig:milc}.

\subsection{Transfer and Supervised Learning}
In the downstream task, we use the representation (output) of the attention model pre-trained using MILC as input to a simple binary classifier on top. Refer to section~\ref{Setup} for further details.

\section{Experiments}
\label{experiments}
In  this  section  we  study  the  performance  of our model on both, synthetic and real data. To compare and show the advantage of pre-training on large unrelated dataset we use three different kind of models --- 1) \phantomsection\label {FPTDef} \FPT\ (Frozen Pre-Trained): The pre-trained model is not further trained on the dataset of downstream task, 2)\phantomsection\label {UFPTDef} \UFPT\ (Unfrozen Pre-Trained): The pre-trained model is further trained on the dataset of downstream task and 3) \phantomsection\label {NPTDef} \NPT\ (Not Pre-trained): The model is not pre-trained at all and only trained on the dataset of downstream task. The models are shown in Figure~\ref{fig:milc}. In each experiment, we compare all three models to demonstrate the effectiveness of unsupervised pre-training.

\subsection{Setup}
\label{Setup}
The CNN Encoder of MILC for simulation experiment consists of $4$ $1$D convolutional layers with output features $(32,64,128,64)$, kernel sizes $(4,4,3,2)$ respectively, followed by ReLU after each layer followed by a linear layer with $256$ units. For real data experiments, we use $3$ $1$D convolutional layers with output features $(64, 128, 200)$, kernel sizes $(4, 4, 3)$ respectively, followed by ReLU after each layer followed by a linear layer with $256$ units. We use stride $1$ for all of the convolution layers.
We also test against autoencoder based pre-training for simulation experiment, for which we use the same CNN encoder as for MILC in the reduction phase. For the decoder, we use the reverse architecture of the encoder that result in $10\times 20$ windows at the output.

In MILC based pre-training, for all possible pairs in the batch, we take feature $z$ from the output layer of CNN encoder. The latent representation of the entire time series is then passed through biLSTM. The output of biLSTM is used as input to the attention model to get a single vector $c$, which represents the entire time series. Scores are calculated using $z$ and $c$ as explained in \ref{milc}. Using these scores, we compute the loss. The neural networks are trained using Adam optimizer.

In downstream tasks we are more interested in subjects for classification task, for each subject the output of attention model ($c$) is used as input to a feed forward network of two linear layers with $200$ and $2$ units to perform binary classification. For experiments, a hold out is selected for testing and is never used through the training/validation phase. For each experiment, 10 trials are performed to ensure random selection of training subjects and, in each case, the performance is evaluated on the hold out (test data). The code is available at: \url{https://github.com/UsmanMahmood27/MILC}


\subsection{Simulation}
To generate synthetic data, we generate multiple $10$-node graphs with $10 \times 10$ stable transition matrices. Using these we generate multivariate time series with autoregressive (VAR) and structural vector autoregressive (SVAR) models~\cite{lutkepohl2005new}.

$50$ VAR times series with size $10 \times 20000$ are split into three time slices respectively for training, validation and testing. Using these samples,
We pre-train MILC to assign windows to respective time series.

In the final downstream task, we classify the whole time-series into VAR or SVAR (obtained by randomly dropping $20\%$  VAR samples) groups.
We generate $2000$ samples and split as $1600$ for training, $200$ for validation and $200$ for hold-out test. For both pre-training and downstream task, we follow the same set up as described in section \ref{Setup}.

 We compare the effectiveness of MILC with the model used in \cite{mahmood2019transfer} and two variations of autoencoder based pre-training. The two variations of autoencoder are acquired by replacing the CNN encoder of \cite{mahmood2019transfer} and MILC by the pre-trained or randomly initialized autoencoder during downstream classification, depending on the model as explained in section \ref{experiments}. We refer to these two variations as \emph{AE\_STDIM} and \emph{AE\_STDIM}+\emph{attention}. Note that difference between the two is the added attention layer in the later during downstream classification.

\begin{figure}
\includegraphics[width=\linewidth, ]{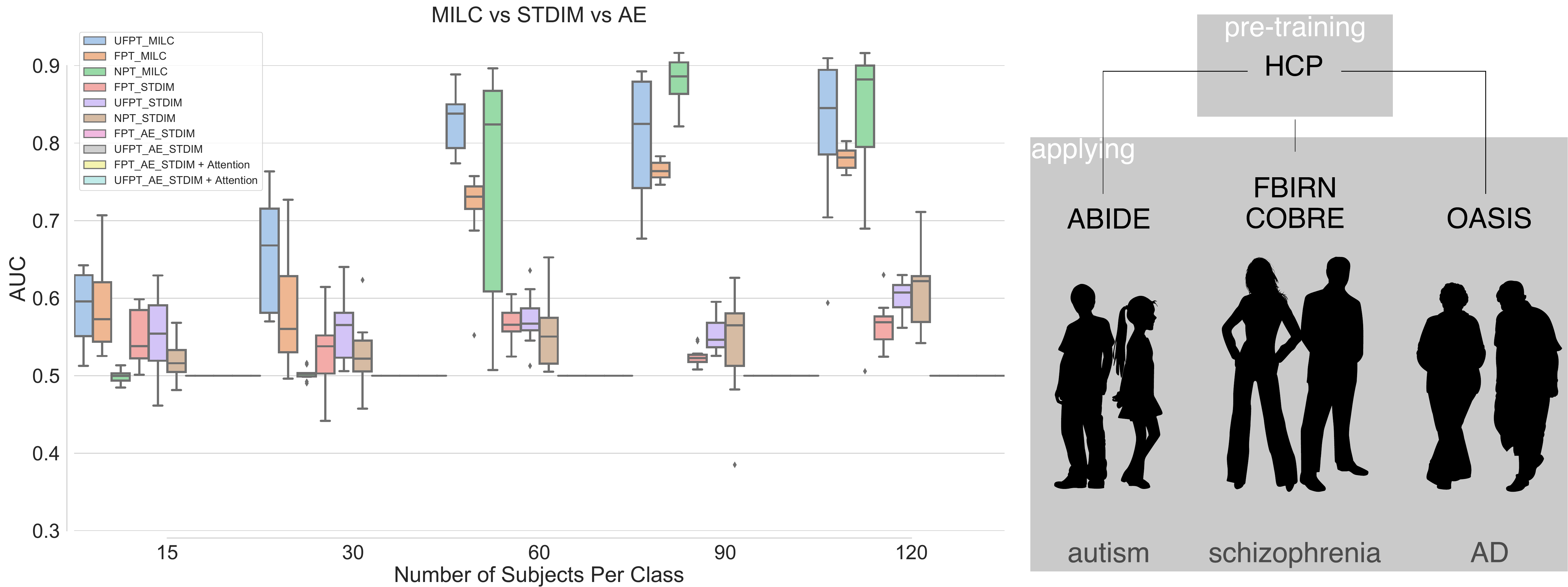}
  \caption[Caption for LOF]{ \textbf{Left:} Area Under Curve (AUC) scores for VAR vs. SVAR time-series classification using MILC, ST-DIM and autoencoder based pre-training methods. MILC based pre-training greatly improves the performance of downstream task with small datasets. On the other side, ST-DIM works better than autoencoder based pre-training which completely fails to learn dynamics and thus exhibits poor performance. \textbf{Right: }  Datasets used for pre-training and classification tasks. Healthy controls from the HCP~\cite{van2013wu} are used for pre-training guided by data dynamics alone\footnotemark[1]. The pre-trained model is then used in downstream classification tasks of $3$ different diseases, $4$ independently collected datasets, many of which contain data from a number of sites, and consist of populations with significant age difference. The age distributions in the datasets have the following mean and standard deviation: \textbf{HCP:} $29.31 \pm 3.67$; \textbf{ABIDE:} $17.04 \pm 7.29$; \textbf{COBRE:} $37.96 \pm 12.90$; \textbf{FBIRN:} $37.87 \pm 11.25$; \textbf{OASIS:} $67.67 \pm 8.92$.
}
  \label{fig:synth_test}
\end{figure}

 It is observed that the MILC based pre-trained models can easily be fine-tuned only with small amount of downstream data. Note, with very few  samples, models based on the pre-trained MILC (\FPT{} and \UFPT{}) outperform the un-pre-trained models (\NPT{}), ST-DIM models, autoencoder based models. ST-DIM based pre-training model \cite{mahmood2019transfer} performs reasonably well compared to autoencoder and \NPT{} models, however, MILC steadily outperforms ST-DIM. Results show that autoencoder based self-supervised pre-training does not assist in VAR vs. SVAR classification. Refer to Figure \ref{fig:synth_test} {\bf Left} for the results of simulation experiments.

\subsection{Brain Imaging}
\label{RealData}
\subsubsection{Datasets}
Next, we apply MILC to brain imagining data. We use rsfMRI data for all brain data experiments. Refer to Figure~\ref{fig:synth_test} for the details of the datasets used. We compare MILC with ST-DIM based pre-training shown in \cite{mahmood2019transfer}.

Four datasets used in this study are collected from \footnotetext[1]{Human silhouettes are by Natasha Sinegina for
  Creazilla.com without modifications,
  \href{https://creativecommons.org/licenses/by/4.0/}{https://creativecommons.org/licenses/by/4.0/}}
FBIRN (Function Biomedical Informatics Research Network  \footnote[2]{These data were downloaded from Function BIRN Data Repository, Project Accession Number 2007-BDR-6UHZ1.})~\cite{keator2016function} project, from COBRE (Center of Biomedical Research Excellence)~\cite{ccetin2014thalamus} project, from release 1.0 of ABIDE (Autism Brain Imaging Data Exchange \footnote[3] {http://fcon\_1000.projects.nitrc.org/indi/abide/})~\cite{di2014autism} and from release 3.0 of OASIS (Open Access Series of Imaging Studies \footnote[4]{https://www.oasis-brains.org/})~\cite{rubin1998prospective}.

\subsubsection{Preprocessing}
We preprocess the fMRI data using statistical parametric mapping (SPM12, http://www.fil.ion.ucl.ac.uk/spm/) under MATLAB 2016 environment.
After the preprocessing, subjects were included in the analysis if the
subjects have head motion $\le 3^\circ$ and $\le 3$ mm, and with functional data providing near full brain successful normalization~\cite{fu2019altered}.

For each dataset, $100$ ICA components are acquired using the same procedure described in~\cite{fu2019altered}. However, only $53$ non-noise components as determined per slice (time point) are used in all experiments. For all experiments, the fMRI sequence is divided into overlapping windows of $20$ time points with $50\%$ overlap along time dimension.

\subsubsection{Schizophrenia}
For schizophrenia classification, we conduct experiments on two different datasets, FBIRN~\cite{keator2016function} and COBRE~\cite{ccetin2014thalamus}. The datasets contain labeled Schizophrenia (SZ) and Healthy Control (HC) subjects.

\begin{figure}
  \includegraphics[width=\linewidth]{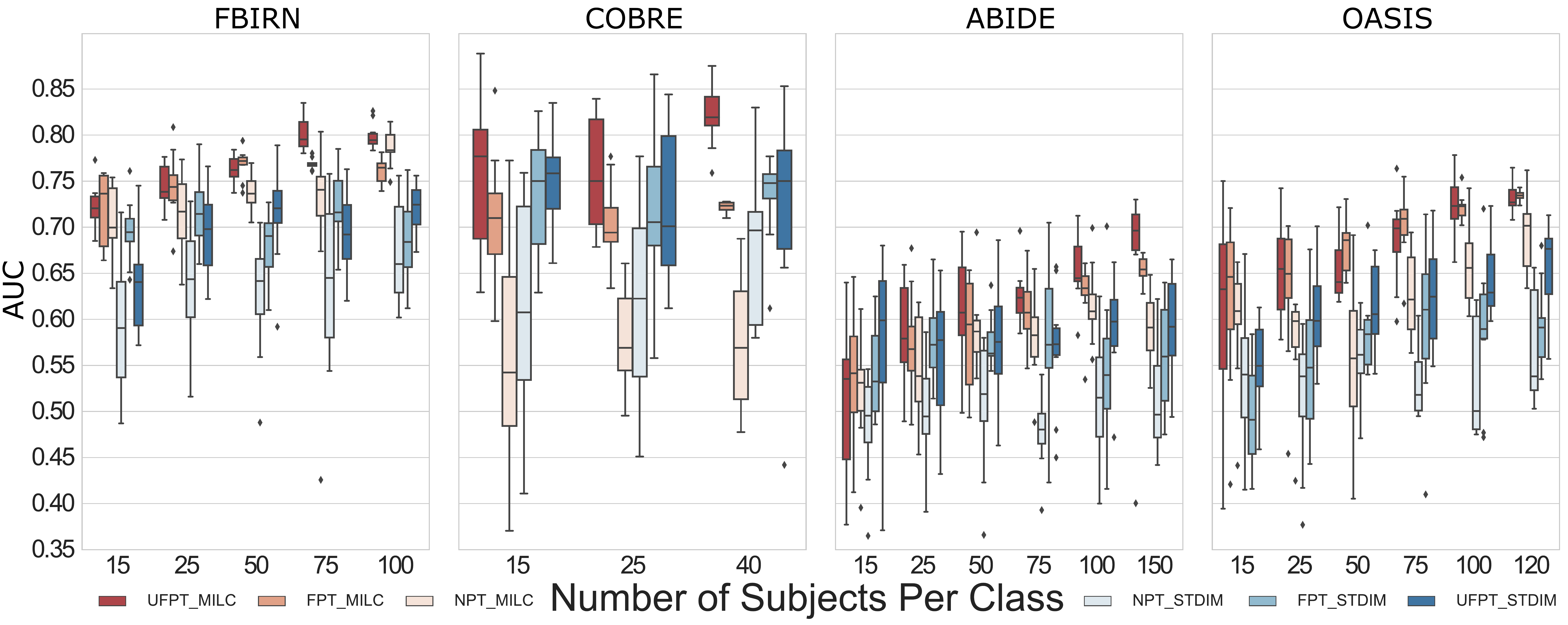}
  \centering
  \caption{AUC scores for all the three models (Refer to Figure \ref{fig:milc}) on real dataset. With every dataset, models pre-trained with MILC (\FPT, \UFPT)  perform noticeably better than not pre-trained model (\NPT{}). Results also show that the learnability of MILC model dramatically increases with small increase in training data (x\_axis). As we can see across the datasets, MILC outperforms ST-DIM with a large margin offering $\sim10\%$ higher AUC when maximum achievable AUC scores are compared.  }
  \label{fig:real_test_AUC_}
\end{figure}

\paragraph{FBIRN}
 The dataset has total $311$ subjects. We use two hold-out sets with sizes $32$ and $64$ for validation and test respectively, remaining are used for supervised training. The details of the results are shown in Figure~\ref{fig:real_test_AUC_}. We see, the pre-trained MILC models outperform \NPT{} and also ST-DIM based pre-trained models.

\paragraph{ COBRE}
 The dataset has total $157$ subjects --- a collection of $68$ HC and $89$ affected with SZ. We use two hold-out sets of size $32$ each for validation and test respectively. The remaining data is used for supervised training. The results in Figure~\ref{fig:real_test_AUC_} strengthen the efficiency of MILC. That is, with only $15$ training subjects, \FPT{} and \UFPT{} perform significantly better than \NPT{} having $\simeq 0.20$ difference in their median AUC scores.

\subsubsection{Autism}
With $569$ total subjects, $255$ are HC and $314$ are affected with autism. We use $100$ subjects each for validation and test purpose. The remaining data is used for downstream training i.e., autism vs. HC classification. Figure~\ref{fig:real_test_AUC_} shows, MILC pre-trained models perform reasonably better than \NPT{} and thus reinforces our hypothesis that unsupervised pre-training learns signal dynamics useful for downstream tasks. We suspect that the reason why pre-trained models do not work well for $15$ subjects is that the dataset is much different than HCP. The big age gap between subjects of HCP and ABIDE is a major difference and $15$ subjects are not enough even for pre-trained models. Refer to Figure~\ref{fig:synth_test} for the demographic information of all the datasets.

\subsubsection{Alzheimer's disease}
The dataset OASIS~\cite{rubin1998prospective} has total $372$ subjects with equal number ($186$) of HC and AZ patients. We use two hold-out sets each of size $64$ respectively for validation and test purpose. The remaining are used for supervised training. Refer to Figure~\ref{fig:real_test_AUC_} for results. The AUC scores of pre-trained models is higher than \NPT{} starting from $15$ subjects, even with $120$ subjects \NPT does not perform equally well.

\subsection{Saliency}
\begin{figure}
  \includegraphics[width=\linewidth]{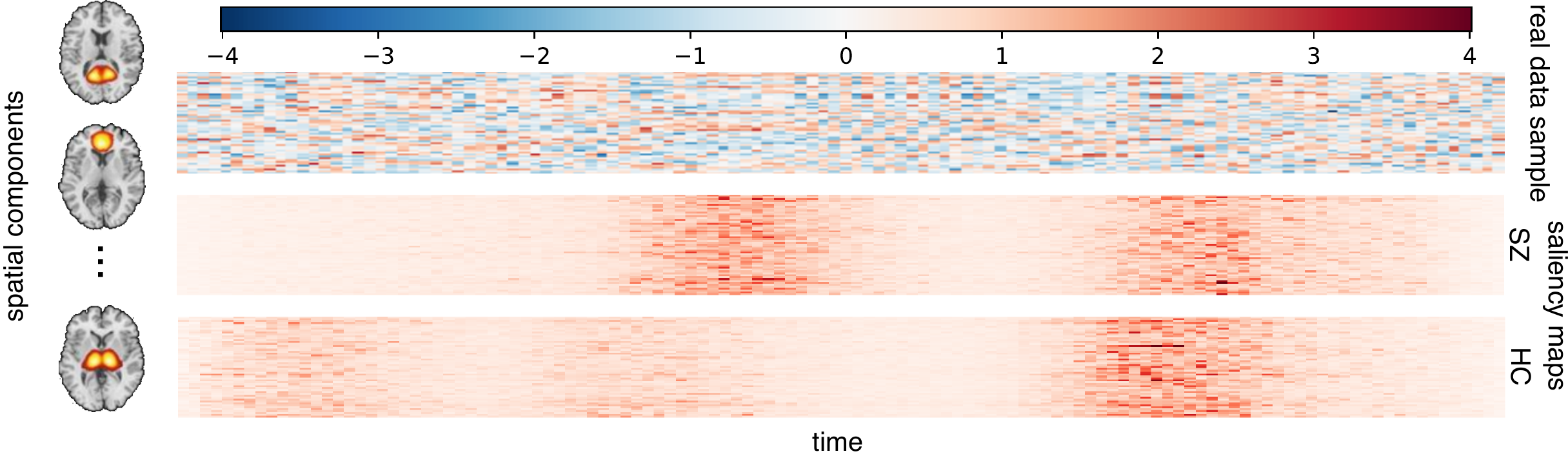}
  \centering
  \caption{Example saliency maps from a pre-trained MILC model: one for a healthy control and one for a schizophrenia subject (FBIRN data). More work is needed, but we already see that not only our model predicts diagnosis but also can point out when during the resting state scan discriminative activity was observed.}
  \label{fig:saliency}
\end{figure}

Our experiments demonstrate that with the whole MILC pre-training we're able to achieve reasonable prediction performance from complete dynamics even on small data. Importantly, we're now able to investigate what in the dynamics was the most discriminative (see Figure~\ref{fig:saliency}).

\section{Conclusions and Future Work}
As we have demonstrated, self-supervised pre-training of a spatio-temporal encoder gives significant improvement on the downstream tasks in brain imaging datasets.
Learning dynamics of fMRI helps to improve classification results for all three dieseases and speed up the convergence of the algorithm on small datasets, that otherwise do not provide reliable generalizations. Although the utility of these results is highly promising by itself, we conjecture that direct application to spatio-temporal data will warrant benefits beyond improved classification accuracy in the future work. Working with ICA components is a smaller and thus easier to handle space that exhibits all dynamics of the signal, in future we will move beyond ICA pre-processing and work with fMRI data directly.
We expect further model introspection to yield insight into the spatio-temporal biomarkers of schizophrenia.
It may indeed be learning crucial information about dynamics that might contain important clues into the nature of mental disorders.

\section{Acknowledgement}
This study was in part supported by NIH grants 1R01AG063153 and 2R01EB006841. We'd like to thank and acknowledge the open access data platforms and data sources that were used for this work, including: Human Connectome Project (HCP), Open Access Series of Imaging Studies (OASIS), Autism Brain Imaging Data Exchange (ABIDE I), Function Biomedical Informatics Research Network (FBIRN) and Centers of Biomedical Research Excellence (COBRE).

\bibliographystyle{splncs04}
\bibliography{paper2938}

\end{document}